\documentclass[twoside]{article}

\usepackage[preprint]{aistats2026}

\usepackage[dvipsnames]{xcolor}
\usepackage{amsmath}
\usepackage{amssymb}
\usepackage{graphicx}
\usepackage{amsthm}
\usepackage{hyperref}
\usepackage[square,numbers]{natbib}
\usepackage{caption}
\usepackage[justification=raggedright]{subcaption}
\usepackage{tikz}
\usetikzlibrary{bayesnet}
\usepackage{xfrac}
\usepackage{cleveref}
\usepackage{nicefrac}
\newtheorem{assumption}{Assumption}
\newtheorem{proposition}{Proposition}
\newtheorem{theorem}{Theorem}

\newtheorem{lemma}{Lemma}

\usepackage[symbol]{footmisc}

\newcommand{\pr}{\mathrm{p}}
\newcommand{\Xc}{\mathcal{X}}
\newcommand{\Rb}{\mathbb{R}}
\newcommand{\Eb}{\mathbb{E}}
\newcommand{\argmax}{\mathrm{argmax}}
\newcommand{\argmin}{\mathrm{argmin}}
\newcommand{\Dc}{\mathcal{D}}
\newcommand{\Lc}{\mathcal{L}}
\newcommand{\Ic}{\mathcal{I}}

\begin{document}

\runningtitle{Learning on Large Scale Screens}

\twocolumn[

\aistatstitle{Accelerated Learning on Large Scale Screens using\\ Generative Library Models}

\aistatsauthor{ Eli N. Weinstein\textsuperscript{$*$,1,2}\And Andrei Slabodkin\textsuperscript{$*$,1} \And Mattia G. Gollub\textsuperscript{$*$,1} \And 
Elizabeth B. Wood\textsuperscript{1,$\dagger$}}

\aistatsaddress{ 1. JURA Bio, Boston, MA, USA \And  2. Technical University of Denmark, Lyngby, Denmark.} ]

\begin{abstract}
Biological machine learning is often bottlenecked by a lack of scaled data.
One promising route to relieving data bottlenecks is through high throughput screens, which can experimentally test the activity of $10^6-10^{12}$ protein sequences in parallel.
In this article, we introduce algorithms to optimize high throughput screens for data creation and model training.
We focus on the large scale regime, where dataset sizes are limited by the cost of measurement and sequencing. We show that when active sequences are rare, we maximize information gain if we \textit{only} collect positive examples of active sequences, i.e. $x$ with $y>0$.
We can correct for the missing negative examples using a generative model of the library, producing a consistent and efficient estimate of the true $\pr(y\mid x)$.
We demonstrate this approach in simulation and on a large scale screen of antibodies.
Overall, co-design of experiments and inference lets us accelerate learning dramatically. 
\end{abstract}

\footnotetext[1]{These authors contributed equally.}
\footnotetext[2]{Contact: \url{ew@jurabio.com}}

Machine learning holds dramatic potential for advancing biological discovery, with growing success in designing proteins, diagnosing genetic disease, predicting pathogen evolution and more \citep{Watson2023-sp,Frazer2021-uq,Thadani2023-hc}. But these and other potential applications are often bottlenecked by a lack of large scale, high quality training data. In particular, there is insufficient information about biological sequences' activity and function.

One promising route to relieving these data bottlenecks is through large scale experiments. Biology provides tools to synthesize sequences, deliver them into cells, and measure their functional activity. Large scale screens can test $10^6-10^{12}$ cells in parallel, each with a different sequence design.  The result is a dataset relating sequences, $x$, to their activity, $y$. 
The challenge is that with large numbers of cells, dataset sizes become limited by the cost of measurement, i.e. the cost of recovering $x$ and $y$ from individual cells.

In this article, we introduce methods to optimize large scale screens for dataset creation and model training. 
Rather than measure a random sample from the population of cells, we propose to focus a limited experimental budget on maximally informative measurements. 
We use the theory of Bayesian experimental design to analyze tradeoffs and optimal resource allocation. We develop a measurement strategy and inference approach that can accelerate model training dramatically, extracting orders of magnitude more information than standard approaches. The technique is particularly well-suited to the hardest learning and design problems, where sequences with a desired activity are exceedingly rare.

\section{Problem Setup and Approach}

\begin{figure*}[t]
    \centering
    \includegraphics[width=\linewidth]{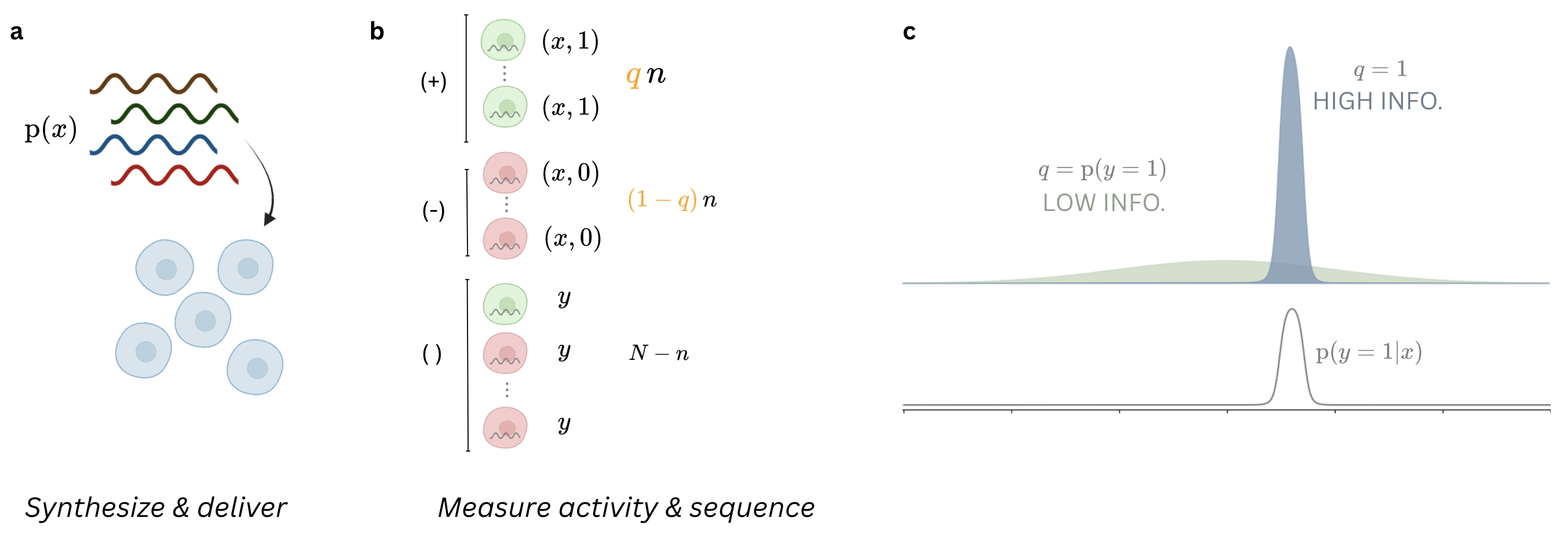}
    \caption{\textbf{Overview of LeaVS.} 
    a. In a large scale screen, sequences $x$ are delivered into cells for testing activity $y$. Each cell receives a sample $x \sim \pr(x)$.
    b. We can measure $(x,y)$ for some cells, but only observe $y$ for the rest. We can choose the fraction $q$ of measurements to allocate to positive and negative examples.
    c. If $\pr(y\mid x)$ is sparse and we collect a representative sample from the library ($q \approx \pr(y=1)$), most $x$ will not be very informative. Allocating all measurements to positive examples ($q=1$) yields more information about $\pr(y\mid x)$. }
    \label{fig:overview}
\end{figure*}

Our goal is to learn a mapping from biological sequences $x$ to functional activity $y$. To do so, we will synthesize different sequences $x$ as DNA, express them as protein, and measure their activity $y$. Then we will train a model that predicts $y$ from $x$. The question is how to accomplish this data generation and model training at scale, such that we obtain a high quality predictor at the end. 
We consider the following experimental approach (\Cref{fig:overview}a).
\begin{enumerate}
\item Synthesize sequences $x$ from a distribution, $\pr(x)$, where $\pr(x)$ is specified by some generative sequence model. (This can be done using variational synthesis \citep{Weinstein2022-sw,Weinstein2024-wm}.)
\item Deliver these sequences, at random, into $N$ different cells. (E.g. using viral vectors.)
\item Measure sequences $x_i$ and activity $y_i$ across cells $i$. (To simplify the discussion, we assume that $y$ is binary, with $y=1$ denoting an active sequence according to some arbitrary threshold.)
\end{enumerate}
The result is a training dataset of x-y pairs, $\{(x_1, y_1), ..., (x_n, y_n)\}$. We can use this data to fit a model of $\pr(y\mid x)$, such as a neural network. 

This experimental approach can reach very large scale in practice. Variational synthesis can create quadrillions of DNA sequences, each an independent sample from $\pr(x)$. Individual sequences can then be delivered into tens of millions of different cells or more. 

In this large scale experimental setting, dataset sizes become limited by the cost of \emph{measurement}, i.e. recovering the x-y pair from each cell. For example, with single cell sequencing it can be intractable to collect data from more than 10-100 thousand cells. 
We study efficient learning algorithms and optimal resource allocation in this measurement-limited regime.

\paragraph{Baseline} A straightforward approach to collecting data would be to measure a random sample of cells from the population. This provides a dataset of i.i.d. samples from the population distribution, $(x_i, y_i) \overset{iid}{\sim} \pr(x, y)$ for $i \in \{1, \ldots, n\}$. We can train models on this data using standard techniques, e.g. optimizing a cross-entropy loss.

\paragraph{(Re)allocating measurement} We propose an alternative. We begin from the observation that we typically have some experimental control over which cells we measure, as well as some side information about those cells we do not measure. Many high throughput screening technologies physically separate the initial set of $N$ cells into pools of active and inactive cells, e.g. via a cell sorter (FACS) or pulldown, before sequencing. This has two important practical consequences. 

First, we can decide how many active cells to sequence versus how many inactive cells (\Cref{fig:overview}b). That is, if we have a budget for sequencing $n$ cells, we can decide the fraction $q$ to collect with $y=1$ versus $y=0$. Then we obtain a dataset $\{(x_1, 1), ..., (x_{nq}, 1), (x_{nq+1}, 0), ..., (x_{n}, 0)\}$, where $x_i \sim \pr(x \mid  y_i)$ and $q  \in [0, 1]$ is the chosen parameter.

Second, we know the number of cells that are active and inactive. This implies that we have a dataset of $y$ values for the cells we do not measure, $\{y_{n+1}, \ldots, y_N\}$, though their $x$ values are missing.

In short, revealing the hidden $x_i$ is costly, but we can choose how to allocate our measurement budget to active vs. inactive sequences.

\paragraph{Contributions} First, we introduce a training algorithm that can learn the true sequence-to-activity map $\pr(y\mid x)$ under any $q\in [0, 1]$ (\Cref{sec:method}). We provide evaluations to check the accuracy and calibration of the learned model. Second, we show that for biologically plausible conditional distributions $\pr(y\mid x)$, we should measure only active sequences. That is, to maximize the information we obtain about $\pr(y\mid x)$ for a budget $n$, we should set $q=1$ (\Cref{sec:theory}). We demonstrate on simulated and real datasets, with an application to therapeutic antibody discovery.
We call the combined experimental design and training approach \emph{LeaVS} (Learning from Variational Synthesis), since it rests on the synthesis distribution $\pr(x)$.

\section{Related Work}

We work in the framework of Bayesian experimental design, aiming to maximize the amount of information we collect for a given budget \citep{Lindley1956-ig,Chaloner1995-hv,Rainforth2024-cn}. We are interested in a setting where prior information is weak and we must make a single experimental design choice before seeing any data, rather than sequential experimentation. We use asymptotic theory to derive a general decision rule \citep{Zemplenyi2023-cd}.

As in active learning, we are focused on prediction \citep{Balcan2016-lb,Gal2017-rp,Smith2023-dt}. However, standard active learning chooses points $x$ at which to measure $y$. Here, we have a different experimental design choice, which is about where to measure $x$ rather than $y$. 

Our learning procedure builds on ideas for handling missing data in supervised (deep) learning \citep{Josse2019-yq,Ipsen2022-cq}. We marginalize out uncertainty in $x$. Unusually, many datapoints are missing $x$ in its entirety, not just some entries.

Also related is work on learning from positive and unlabeled data (PU learning). Often, PU learning assumes all the $x$ values are observed, but some of the $y$ values are missing~\citep{Elkan2008-ok,Song2021-rx}. 
Our situation is the opposite: all the $y$ values are observed, but there are missing $x$. More closely related is \citet{Kiryo2017-ap,Plessis2015-ks,Wu2021-fx}, which assume access to samples from $\pr(x)$ and from $\pr(x\mid y=1)$, along with knowledge of $\pr(y=1)$. They use a modified empirical risk minimization objective to train a classifier. We make similar assumptions when $q=1$, but learn using the data's likelihood. This enables Bayesian experimental design, and allows the method to be extended to non-binary $y$.

\citep{Song2021-rx} develop PU methods for learning sequence-activity maps from screens. Their method is designed for smaller scale screens, where we deterministically design, synthesize and test each $x_i$, meaning each $x_i$ is known.
We consider larger scale screens, where sequences must be synthesized stochastically, and each cell receives a random sample from the distribution. In this case, many $x_i$ are latent. 

The goal of large scale screens is often to find active sequences, rather than to learn a model of $\pr(y\mid x)$. Here, it is already common practice to only collect positive examples. Indeed, for assays based on selection, the negative examples are destroyed. Despite common concerns that this data is not suitable for machine learning, we show that it is actually \textit{optimal} in a specific sense.

\section{Method} \label{sec:method}

\subsection{Training}

Our goal is to learn a model that predicts $y$ from $x$, regardless of $q$. To do so, we consider the joint likelihood of all the available data $\mathcal{D}$, including not just x-y pairs, $\mathcal{D}_{xy}$, but also the $y$ values with missing $x$, $\mathcal{D}_y$. Let $\pr_\theta(y\mid x)$ denote a model parameterized by $\theta \in \Theta$, such as a neural network. The data likelihood $\log \pr_\theta(\mathcal{D})$ is
\begin{align} \label{eqn:leavs}
&{\color{blue} \sum_{i=1}^n \log \pr_\theta(y_i \mid  x_i)} + {\color{ForestGreen} \sum_{i=n+1}^N \log \int \pr_\theta(y_i \mid  x) \pr(x) dx} \\ & \triangleq {\color{blue} \mathcal{L}_{xy}(\theta)} + {\color{ForestGreen} \mathcal{L}_{y}(\theta)}
\end{align}
The first term (in blue) is the standard regression likelihood; for binary $y$, it is the cross-entropy. The second term (in green) is the marginal likelihood of $y_i$. Since $x_i$ is missing, we integrate over its distribution.

How can we compute $\mathcal{L}_{y}(\theta)$ in practice? We rely on access to samples from $\pr(x)$. In particular, if we are using variational synthesis to produce a large library, we have a generative model specifying the library distribution $\pr(x)$ \citep{Weinstein2022-sw}. 
We draw samples, $x'_1, ..., x'_M \sim \pr(x)$ and compute,
\begin{equation} \label{eqn:mc-approx}
\int \pr_\theta(y_i \mid  x) \pr(x) dx  \approx \frac{1}{M} \sum_{j=1}^M \pr_\theta(y_i \mid  x'_i).
\end{equation}  
For optimization, we can differentiate the approximation with respect to $\theta$, as $\pr(x)$ is independent of $\theta$. 

We refer to the log likelihood in \Cref{eqn:leavs} as the LeaVS (Learning from Variational Synthesis) objective, since it incorporates the synthesis model $\pr(x)$. Intuitively, the LeaVS objective is akin to data augmentation: we augment our training dataset with $x$ that are computationally generated. Unlike standard data augmentation, however, we do not give these generated sequences fixed labels, but rather use them to approximate an integral.

\subsection{Learning under Alternative Allocations} 

The key property of the LeaVS objective is that it allows us to learn $\pr(y \mid x)$ under different measurement allocations $q$. In particular, we propose to set $q=1$.
Here, we summarize the results in \Cref{sec:theory}. 

First consider just using data on x-y pairs, $\mathcal{D}_{xy}$, and learning a model based on the likelihood $\mathcal{L}_{xy}(\theta)$. Then we will only learn the correct predictor $\pr(y \mid  x)$ if we measure a representative sample from the library, i.e. we must set $q = \pr(y=1)$ to obtain a consistent estimate of $\pr(y \mid  x)$ (\Cref{thm:inconsistent}).

With the LeaVS objective $\mathcal{L}_{xy}(\theta) + \mathcal{L}_{y}(\theta)$ the situation changes markedly. We can now learn the correct predictor \emph{regardless} of how we allocate measurement resources.
That is, $\pr_\theta(y \mid x)$ will converge to the true $\pr(y \mid x)$ for any $q \in [0,1]$ (\Cref{thm:consistent}).
Notably, this range includes $q = 1$: measuring \textit{only} positive examples. The LeaVS objective automatically accounts for the unobserved inactive sequences, through knowledge of the underlying library distribution $\pr(x)$.

In general, the optimal choice of $q$ depends on the situation. But biological sequence-to-activity maps show a recurring feature: only sequences within a narrow region of sequence space $\Xc$ show non-trivial activity (\Cref{fig:overview}c). So, positive examples provide a large amount of information: they pin down the region's location quite precisely. Negative examples, on the other hand, tell us little about the region's location.
As a result, focusing our measurements on positive examples maximizes information gain (\Cref{thm:argmax_1}).

In summary, we propose to change our experimental design (set $q=1$) and change our training objective (add $\mathcal{L}_{y}(\theta)$) to accelerate learning of $\pr(y\mid x)$.

\subsection{Model evaluation}

We have shown that it is possible to learn the sequence-to-activity map under different measurement allocations. But for reallocation to be practical, we must also be able to evaluate the models we train, to check their performance. This is non-trivial if we only have access to positive examples, $q=1$.

\paragraph{Accuracy} To evaluate accuracy based on positive examples, we again make use of the library distribution $\pr(x)$. Let $t_\theta(x) = \mathbb{I}(\pr_\theta(y|x) > 0.5)$ be a classifier based on $\pr_\theta(y | x)$. Applying the approach of \citet{Kiryo2017-ap}, we decompose the accuracy $\textsc{A}_\pr[t_\theta] = \mathbb{E}_\pr[Y t_\theta(X)] + \mathbb{E}_\pr[(1-Y) (1 - t_\theta(X))]$ as
\begin{align*}
\textsc{A}_\pr[t_\theta] =& \pr(y=1) \mathbb{E}[t_\theta(X)|Y=1] + \mathbb{E}_\pr[1 - t_\theta(X)]\\ & - \pr(y=1) \mathbb{E}_p[1 - t_\theta(X)|Y=1]
\end{align*}
The first term is the fraction of predictions that are true positives; the second is the fraction of negative predictions; the last is the fraction of false negatives. The first and third terms just depend on $\pr(y)$ and $\pr(x \mid y=1)$, so can be approximated using heldout data. The second term depends on $\pr(x)$, so can be approximated using the library model (\Cref{apx:accuracy}).

We can apply the same strategy to approximate false positive and false negative rates, or ROC and precision-recall curves, by plugging different thresholds into $t_\theta(x)= \mathbb{I}(\pr_\theta(y|x) > 0.5)$ (\Cref{apx:precision-recall}).

\paragraph{Calibration} We would like models that are not only accurate, but also quantify their uncertainty. For example, many algorithms for designing and optimizing sequences require upper confidence bounds. A standard evaluation metric is the expected calibration error (ECE)~\citep{Vaicenavicius2019-ir}. 

The ECE asks whether, if the model says we should see $Y = 1$ some percentage of the time, then we actually see $Y = 1$ at that percentage. $\textsc{ece}_\pr[\pr_{\theta}]$ is
\begin{equation}
	\mathbb{E}_{\pr}\left[d\left(\pr\left(y=1 \mid \pr_{\theta}(y=1|X)\right), \pr_{\theta}(y=1|X)\right)\right]
\end{equation} 
Here $d(\cdot, \cdot)$ is a distance function, such as the absolute difference.
It is standard to use a histogram based estimator for the ECE.
We define bins $\Phi_1, \ldots, \Phi_K$, where $\Phi_1 = [0, 1/K), \Phi_2 = [1/K, 2/K), \ldots$, then evaluate,
\begin{equation*}
    d\left( \mathbb{E}_{\pr}[Y \mid \pr_{\theta}(y=1|X) \in \Phi_k], \mathbb{E}_{\pr}[\pr_{\theta}(y=1|X) \mid \pr_{\theta}(y=1|X) \in \Phi_k]\right)
\end{equation*}
We can estimate $\mathbb{E}_{\pr}[Y \mid \pr_{\theta}(y=1\mid X) \in \Phi_k] $ as,
\begin{equation*}
    \frac{\pr(Y=1) \mathbb{E}_{\pr}[ \mathbb{I}(\pr_\theta(y=1\mid X) \in \Phi_k) \mid Y=1]}{\mathbb{E}_{\pr}[\mathbb{I}(\pr_\theta(y=1\mid X) \in \Phi_k)]},
\end{equation*}
where again we find only quantities that depend on $\pr(y=1)$, $\pr(x)$ and $\pr(x \mid y=1)$. Details in \Cref{apx:calibration}.

In summary, it is possible to evaluate models using heldout positive examples. So we can critique and improve our models even when $q=1$.

\section{Theory} \label{sec:theory}

We have proposed methods to train and evaluate predictive models under different measurement allocations, $q \in [0, 1]$. 
In this section, we ask which $q$ to choose in practice.
We consider a class of distributions that is widely found in biology. 
For this class, $q=1$ maximizes the information we obtain about $\pr(y\mid x)$.

We work in the framework of Bayesian experimental design~\citep{Lindley1956-ig,Chaloner1995-hv,Rainforth2024-cn}.
The goal is to choose a design that maximizes the information we will gain from collecting data.
Here, the experimental design parameter is $q$. Based on $q$, we obtain a (random) dataset $\Dc_q$. After observing this data, we obtain a posterior over the unknown parameter $\theta$. Its Shannon entropy is,
\begin{equation}
	\mathcal{H}(\pr(\theta \mid \mathcal{D}_q)) = - \int  \pr(\theta \mid \mathcal{D}_q) \log \pr(\theta \mid \mathcal{D}_q):
\end{equation}
Our goal is to set $q$ such that the entropy will be minimized, indicating that we have learned as much as possible about $\pr(y \mid x)$:
\begin{equation}
	\underset{q \in [0, 1]}{\argmin}\, \mathcal{H}(\pr(\theta \mid \mathcal{D}_q)).
\end{equation}

\subsection{Sparse Activity}

In general, the optimal choice of $q$ will depend on the data distribution $\pr(x, y)$ and the model $\pr_\theta(y \mid x)$. We will focus on a class of distributions that are common in biological applications.

It is often observed that only sequences in a small region of sequence space have any biological function or activity. 
That is, activity is \emph{sparse}.
To formalize this mathematically, note we can write any $\pr_\theta(y \mid x)$ as,
\begin{equation} \label{eqn:sparse_active}
	\pr_\theta(y=1 \mid x) = \begin{cases}
		h_\theta(x) & x \in S_\theta\\
		0 & x \notin S_\theta
	\end{cases}
\end{equation} 
for $\theta \in \Theta \subseteq \Rb^d$. Here $S_\theta \subseteq \Xc$ defines the region of sequence space where the model predicts non-zero activity.
We will assume that the true data distribution falls into the model class,
\begin{assumption}[Well-specified model] \label{asm:specified}
	There exists $\theta_0 \in \Theta$ such that $\pr_{\theta_0}(y \mid x) = \pr(y \mid x)$.
\end{assumption}
With this class of distributions, $\eta \triangleq \pr(x \in S_{\theta_0})$ gives the probability of a sequence falling in the active region. $\eta$ is a measure of how rare active sequences are across the sequence landscape. 
We are interested in the regime where $\eta$ is small, i.e., activity is sparse. 
This is especially common in the most challenging molecular discovery problems.

\subsection{Asymptotic consistency}

We are interested in the large scale screening regime, where we are collecting big datasets that are limited in size by the cost of measuring $x$. 
To study this regime, we first take $N \to \infty$, so that there is an excess of y-only data. Then, we examine the behavior of the posterior as $n \to \infty$, i.e. as we collect more x-y pairs.

When $N \to \infty$, the likelihood contribution $\Lc_y(\theta)$ dominates \Cref{eqn:leavs}. Asymptotically, $\Lc_y(\theta)$ will be maximized when $\int \pr_\theta(y\mid x) \pr(x) dx = \pr(y)$, i.e. if the model captures the true frequency of positive and negative examples.
So to study the $x$ measurement-limited regime, we make the simplifying assumption,
\begin{assumption}[Known hit rate] \label{asm:known_hit}
	$\int \pr_\theta(y \mid x) \pr(x) = \pr(y = 1)$ for all $\theta \in \Theta$.
\end{assumption}

We are now interested in understanding the behavior of the posterior as function of $n$, the number of x-y pairs we observe. After setting $q$, we observe,
\begin{align}
	X_{1, \ldots, n q} \overset{iid}{\sim} \pr(x \mid y = 1)\\
	X_{n q+1, \ldots, n} \overset{iid}{\sim} \pr(x \mid y = 0),
\end{align}
and we are interested in the posterior,
\begin{equation}
	\pr(\theta \mid \mathcal{D}_{n,q}) \propto \pr(\theta)\prod_{i=1}^{nq} \pr_\theta(y_i=1 \mid x_i)\prod_{i=nq +1}^{n} \pr_\theta(y_i=0 \mid x_i),
\end{equation}
under the constraint of \Cref{asm:known_hit}.

We first show that this posterior concentrates at the true $\theta_0$ regardless of the choice of $q$.
We assume the distribution $\pr(x)$ has full support and the prior covers the true parameter, $\pi(\theta_0) > 0$, and we place standard regularity conditions on the likelihood's parameterization \citep{Miller2021-xj}. Details and proof in \Cref{apx:consistent}. 
\begin{proposition}[Consistent for any $q$] \label{thm:consistent}
	Under \Cref{asm:known_hit}, \Cref{asm:support} and \Cref{asm:BvM}, we have $\pr(\theta \mid \Dc_{n,q}) \to \delta_{\theta_0}$ a.s. as $n \to \infty$, for $q \in [0, 1]$.
\end{proposition}
The y-only data is essential to learning $\theta_0$ under different allocations $q$. If we use a model without the constraint from \Cref{asm:known_hit}, consistency will fail. Details and proof in \Cref{apx:inconsistent}.
\begin{proposition}[Inconsistent without $y$ data] \label{thm:inconsistent}
    Under \Cref{asm:shift-spec} and \Cref{asm:bvm-inconsistent}, we have $\pr(\theta \mid \Dc_{n,q}) \to \delta_{\tilde \theta}$ a.s. as $n \to \infty$, where $\pr_{\tilde \theta}(y \mid x) \neq \pr(y \mid x)$ except at $q = \pr(y=1)$.
\end{proposition}

\subsection{Efficiency and optimal $q$} \label{sec:theory_back_envelope}

We now show that when activity is sparse, posterior information is asymptotically maximized by collecting only positive examples.

First, we look more closely at the behavior of the posterior as we collect more data. The Bernstein-von Mises theorem says the posterior will approach a Gaussian, under regularity conditions on the likelihood. Let $\pr_q(y)$ denote $\mathrm{Bernoulli}(q)$.
\begin{theorem}[Bernstein-von Mises]\citep[e.g.][Thm. 3.2]{Miller2021-xj}. \label{thm:BvM}
Let \Cref{asm:known_hit}, \Cref{asm:support} and \Cref{asm:BvM} hold. Let $r_n(\tilde \theta)$ be the density of $\sqrt{n} (\theta - \theta_n)$, where $\theta_n$ is the maximum likelihood estimate. Then, $\pr(\tilde \theta \mid \Dc_{q,n})$ converges a.s. in total variation to a normal distribution with mean 0 and inverse covariance given by the information matrix,
\begin{equation}
	H_{q} = - \Eb_{\pr(x \mid y)\pr_q(y)}[\nabla^2_\theta \log \pr_\theta(Y \mid X) |_{\theta = \theta_0}]
\end{equation}
\end{theorem}
So, to maximize the amount of information we obtain, we will minimize the entropy of the asymptotic posterior distribution (D-optimality \citep{Chaloner1995-hv,Huan2024-xa}),
\begin{equation}
	\underset{q \in [0, 1]}{\argmin}\, \mathcal{H}(\mathcal{N}(0, H_q^{-1})). \end{equation}

We find that when the size $\eta$ of the active region is sufficiently small, the asymptotic posterior entropy is minimized at $q = 1$. Let $S_0 \triangleq S_{\theta_0}$ and define information matrices over $S_0$,
\begin{align}
	\Ic_y = -\Eb_{\pr(x \mid S_0, y)}[\nabla^2_\theta \log \pr_\theta(Y \mid X) |_{\theta_0}]
\end{align}
for $y \in \{0, 1\}$. Note $\Ic_0$ and $\Ic_1$ can be positive definite because the true $\theta_0$ is identified from $\pr(x \mid y=1)$ or, by symmetry, from $\pr(x \mid y=0)$. 
\begin{proposition} \label{thm:argmax_1}
	Assume $\Ic_0$ and $\Ic_1$ are both positive definite. Then if,
\begin{equation} \label{eqn:condition_q1}
	\det(\Ic_1 - \frac{\eta}{1 - \eta} \Ic_0) > 0
\end{equation}
we have ${\argmax}_q\, \mathcal{H}(\mathcal{N}(0, H_q)) = 1$. This holds for any $\eta$ sufficiently small.
\end{proposition}
Proof in \Cref{apx:argmax_1}. 
The key idea is that, when $\eta$ is small, most of the negative examples will come from outside the active region $S_0$ (\Cref{fig:overview}c). These provide very little information about $\pr(y \mid x)$. Positive examples, meanwhile, provide large amounts of information, since they come from within the active region.
By measuring only positive examples ($q=1$), we concentrate our data collection efforts in the region of sequence space that provides the most information.

When should we set $q=1$ in practice? \emph{A priori} we do not know $\Ic_1$, $\Ic_0$ or $\eta$. For a back-of-the-envelope calculation, assume that within $S_0$, positive and negative examples are roughly equally common and equally informative, such that $\Ic_1 \approx \Ic_0$ and $\eta \approx 2 \pr(y = 1)$. 
Then, if $\pr(y = 1) < 1/4$, \Cref{eqn:condition_q1} will hold.
So, back-of-the-envelope, we should collect only positive examples if we expect hit rates below about 25\%.
In practice, we can often estimate $\pr(y=1)$ before selecting $q$ and measuring $x$, e.g. after sorting but before sequencing, so this decision rule is actionable.

How much information will we gain once we set $q=1$? Back-of-the-envelope, we gain more than $-\frac{d}{2} \log 3\pr(y=1)$ nats, where $d$ is the dimension of the parameter space $\theta$ (\Cref{apx:info_gain}). 
This corresponds to the same information gain, asymptotically, as increasing the sample size from $n$ to $\nicefrac{n}{3 \pr(y=1)}$.
Empirically, in challenging therapeutic discovery problems, $\pr(y=1)$ can be one in a billion \citep{Skora2015-qo}. 
This suggests that in practice, we could increase our effective data set size by a factor of hundreds of millions.

\section{Results}

We examine the empirical performance of LeaVS on both synthetic and experimental data. 
Our key finding is that the right combination of experimental design (measure only positive examples) and training algorithm (the LeaVS objective) yields better predictions.

Our theoretical results are based on an asymptotic analysis of Bayesian learning with a parametric model. In practice, non-Bayesian, deep learning models are widely used for sequence-to-activity maps.
Here, we show empirically that our key theoretical findings carry over to this setting.
We train transformer and CNN-based models, and approximate the LeaVS objective using stochastic minibatching. Details in \Cref{apx:deep-models}.

\begin{figure*}[t]
    \centering
    \begin{subfigure}[t]{0.3\linewidth}
    \centering
    \includegraphics[width=\linewidth]{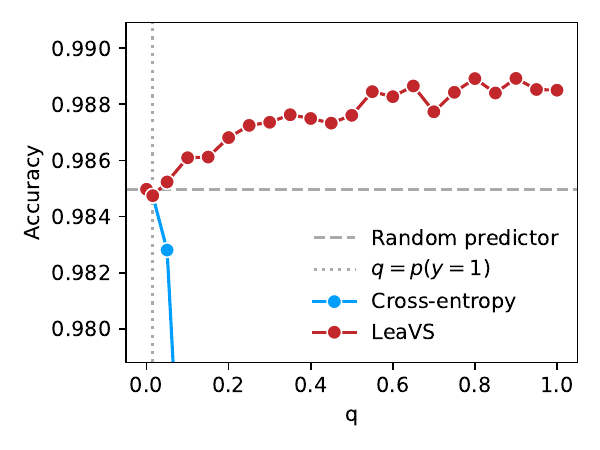}
    \caption{Accuracy.} \label{fig:accuracy_zoomed}
    \end{subfigure}
    \begin{subfigure}[t]{0.4\linewidth}
    \centering
    \includegraphics[width=0.75\linewidth]{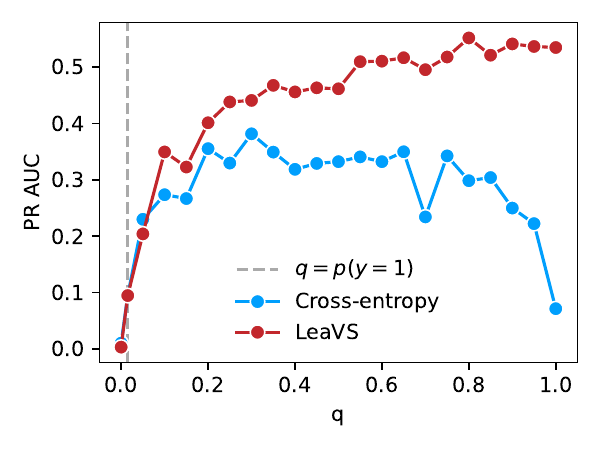}
    \caption{Area under the precision-recall curve.}
    \end{subfigure}
    \caption{\textbf{Performance on synthetic data.} }
    \label{fig:synthetic}
\end{figure*}

\subsection{Synthetic data}
We first examined a synthetic data setting, where the true $\pr(y \mid x)$ is known.
We set $\pr(x)$ to be a variational synthesis model of antibody CDRH3 loops \citep{Weinstein2024-wm}. 
Following previous studies of antibody binding, we set $\pr(y \mid x)$ such that $y$ depends on the presence of specific amino acid motif in $x$ \citep{Akbar2021-mk,Pavlovic2021-sn}.
The overall activity rate was small, at $\pr(y=1) = 0.015$. Details in \Cref{apx:synthetic}. 

We generated datasets under different measurement allocations $q$, and trained transformer-based models using cross-entropy and the LeaVS objective.
Examining model accuracy, we see very different performance as a function of $q$ (\Cref{fig:accuracy_zoomed}). 
Standard cross-entropy peaks around $q=\pr(y=1)$, i.e. when we take a representative sample from the distribution. But even at this peak, it does not surpass random predictions with frequency $\pr(y=1)$.
Training with LeaVS, we find similar performance at $q=\pr(y=1)$. 
But now, as we turn up $q$, the model improves substantially, peaking near $q=1$, even as the cross-entropy model's accuracy drops to near zero (\Cref{fig:accuracy_full}).
In short, by modifying both the experimental design and the training procedure, we achieve substantial performance gains, while either modification on its own is insufficient.

As an additional performance metric, we also considered the area under the precision-recall curve. This metric is commonly used for imbalanced datasets, and is, unlike accuracy, insensitive to rescaling the prediction $\pr_{\hat \theta}(y \mid x)$. 
We find qualitatively similar behavior. 
Cross-entropy peaks at low values of $q$, while LeaVS peaks near $q=1$, achieving much better performance.

We also checked how reliable model evaluation is, in the absence of negative examples. 
We see a close match between estimated and true accuracy, sufficient to choose a high-quality model based on held-out data (\Cref{fig:accuracy_estimated}). Calibration error estimates are similarly reliable (\Cref{fig:calibration}).

\subsection{Experimental data from TCRs} 

We next sought to evaluate LeaVS on real experimental data, where $\pr(x, y)$ is given by nature. 
We used a dataset screening human T cell receptors (TCRs), $x$, for binding against an influenza antigen, $y$ \citep{Genomics2019-ks}.
In this dataset, we have measurements of $(x, y)$ for each cell, obtained via single cell sequencing.
Starting from this complete dataset, we generated smaller datasets with different values of $q$, by subsampling cells.
To obtain samples from $\pr(x)$ for training, we drew samples of $x$ from heldout cells.
Details in \Cref{apx:semi-experimental}.
Note that because we must subsample a full dataset, $n$ is relatively small, 170 cells, which increases variability.

\begin{figure*}[t]
    \centering
    \begin{subfigure}[t]{0.3\linewidth}
    \centering
    \includegraphics[width=\linewidth]{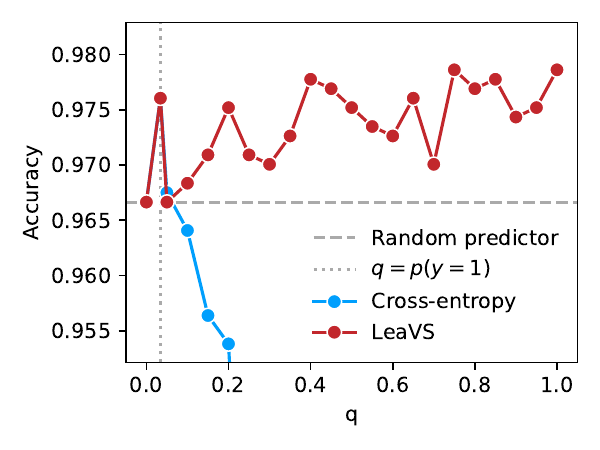}
    \caption{Accuracy.}
    \end{subfigure}
    \begin{subfigure}[t]{0.4\linewidth}
    \centering
    \includegraphics[width=0.75\linewidth]{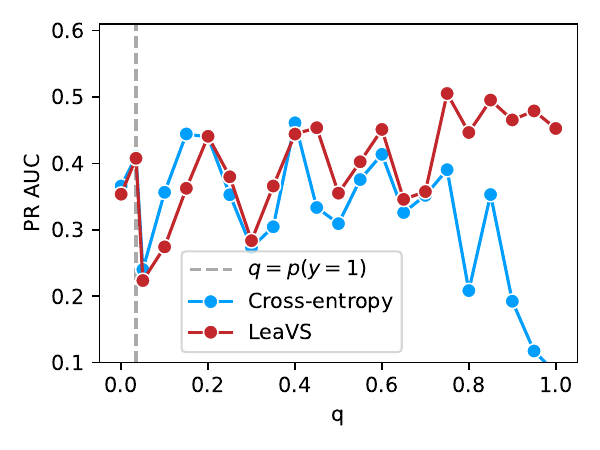}
    \caption{Area under the precision-recall curve.}
    \end{subfigure}
    \caption{\textbf{Performance on TCR data.} }
    \label{fig:semiexperiment}
\end{figure*}

The results on this experimental data, shown in \Cref{fig:semiexperiment}, show a similar pattern to the synthetic data.
Combining high $q$ with the LeaVS objective boosts performance over cross-entropy at $q=\pr(y=1)$. Alone, increasing $q$ or using the LeaVS objective does not help; the experiment and inference must be modified together.
Accuracy and calibration error estimates continue to track the true values, though they are sometimes conservative, under-estimating accuracy and over-estimating calibration (\Cref{fig:semiexperiment-evals}).
Overall, we find the LeaVS methodology produces better models of a real sequence-to-activity relationship.

\subsection{Large scale demonstration on antibodies}

Finally, we deploy LeaVS at large scale, to learn therapeutically important sequence-activity relationships. 
We focus on TCR mimicking antibodies (TCRm) and their binding against a challenging oncology target \citep{Klebanoff2023-bx}.
Previously, we used variational synthesis to synthesize $10^{16}$ samples from a generative model of human antibody CDRH3s. 
The DNA was assembled into scFv CAR cell therapy constructs and delivered into 22.5 million human cells.
The library was screened against a panel of fluorescent and DNA-barcoded peptide-HLA (pHLA) oncology targets.
We obtained $y$ values for the full population of cells based on sorting (FACS). 
Single cell sequencing was used to recover synthesized antibody sequence, $x$, along with counts measuring binding strength, $y$. Given the low value of $\pr(y > 0)$ estimated from sorting (below 1\%), we chose $q=1$ and allocated our entire single cell sequencing budget to active sequences with $y > 0$ (\Cref{sec:theory_back_envelope}). Details in \Cref{apx:vs-demo}.

\begin{figure*}[t]
    \centering
    \begin{subfigure}[t]{0.3\linewidth}
    \centering
    \includegraphics[width=\linewidth]{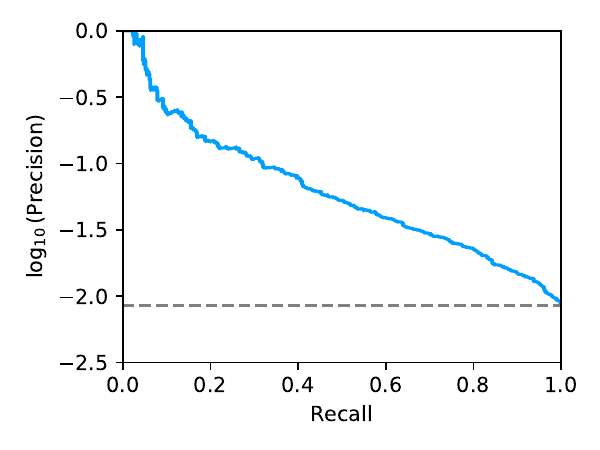}
    \caption{(Log) precision-recall. AUC: 0.12.} \label{fig:pr_curve}
    \end{subfigure}
    \begin{subfigure}[t]{0.34\linewidth}
    \centering
    \includegraphics[width=\linewidth]{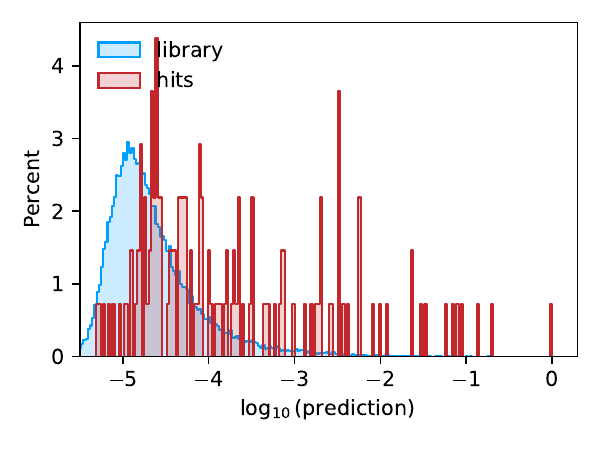}
    \caption{Predictive scores on library sequences (blue) versus heldout positive examples (red). } \label{fig:histograms}
    \end{subfigure}
\caption{\textbf{Predicting binding of TCRm scFv CAR therapeutic candidates  to MAGE-A4, a challenging oncology target.} }
    \label{fig:experiment}
\end{figure*}

Using the LeaVS objective, we trained a model to predict binding to the cancer-testis antigen MAGE-A4.
The LeaVS objective, \Cref{eqn:leavs}, extends to non-binary $y$, so we trained directly on binding counts, though we binarize $y$ for evaluation. 
We used the antibody variational synthesis model for $\pr(x)$.
We find that the LeaVS-trained predictor successfully generalizes to heldout data, surpassing a random predictor (\Cref{fig:pr_curve}).
Indeed, we see precision values more than an order of magnitude larger than the baseline hit rate (dashed), implying that if we select sequences with high values of $\mathbb{E}[Y \mid x]$ we can enrich for MAGE-A4 hits by 10-100x. This suggests the model is capable of being used for iterative sequence design.

To confirm the model's predictive ability was not an artifact of misspecification of $\pr(x)$, we used sequencing data from the pre-screen DNA library.
The model's estimate of $\mathbb{E}[Y \mid x]$ distinguishes this distribution from the distribution of hits (\Cref{fig:histograms} and \Cref{fig:transformer_experiment_sequenced}).
These results are robust to changes in model architecture, e.g. using a CNN in place of a transformer (\Cref{fig:cnn_experiment}).

In short, we can use LeaVS to train therapeutically relevant predictive models, describing interactions between generative model-designed antibodies and challenging oncology targets.

\section{Discussion}

We have proposed LeaVS, a new approach to scaling up data generation for biological machine learning. LeaVS rests on the co-design of experiments and training algorithms. 
It produces more informative datasets by focusing limited measurement budgets on positive examples. 
Then, it adjusts for missing data during training, using a generative model of the experiment.
By shifting the distribution of training data, then correcting for this shift post-hoc, we accelerate learning.
But neither change is useful on its own; we have to consider experiments and training together.

\paragraph{Limitations and assumptions}
A key assumption of the method is that we have access to samples from $\pr(x)$. 
These samples can come from a variational synthesis model of the library - but this model may not be perfect. Alternatively, samples can come from sequencing part of the library before screening - but this is costly, and provides only a limited number of samples. 
Developing methods that are robust to limited samples or misspecification of $\pr(x)$ is an important area for future work.

The optimal $q$ depends on unobserved quantities. 
We have relied on back-of-the-envelope estimates to suggest that we set $q=1$ when we observe $\pr(y=1) < 0.25$ (\Cref{sec:theory_back_envelope}). 
But if, e.g., the sequence-to-activity map is not sparse, this may not be optimal.

\paragraph{Future directions} The LeaVS objective includes a marginal likelihood, which we approximate with a nested Monte Carlo estimate (\Cref{eqn:mc-approx}) \cite{Rainforth2018-qp}. This estimate might be improved using importance sampling, or multilevel Monte Carlo methods \cite{Rainforth2024-cn,Goda2022-pc}.

A major goal of biological machine learning is to design sequences. Modeling $\pr(y \mid x)$ is one route to creating good designs, but other algorithms directly refine a generative model of sequences $\pr(x)$, e.g. through conditioning or reward-based fine-tuning \citep{Abdolmaleki2025-gy}.
It is unclear how best to adapt these algorithms to reap the information gains of setting $q=1$.

More broadly, a fundamental lesson of modern machine learning is that scale is essential to unlocking new model capabilities. 
In biological machine learning, laboratory experiments are a core part of the AI stack. 
Our work emphasizes the importance of considering experiments and training in tandem, to scale up models and accelerate learning.

\bibliographystyle{plainnat}
\bibliography{references}

\begin{thebibliography}{30}
\providecommand{\natexlab}[1]{#1}
\providecommand{\url}[1]{\texttt{#1}}
\expandafter\ifx\csname urlstyle\endcsname\relax
  \providecommand{\doi}[1]{doi: #1}\else
  \providecommand{\doi}{doi: \begingroup \urlstyle{rm}\Url}\fi

\bibitem[Abdolmaleki et~al.(2025)Abdolmaleki, Piot, Shahriari, Springenberg,
  Hertweck, Bloesch, Joshi, Lampe, Oh, Heess, Buchli, and
  Riedmiller]{Abdolmaleki2025-gy}
Abbas Abdolmaleki, Bilal Piot, Bobak Shahriari, Jost~Tobias Springenberg, Tim
  Hertweck, Michael Bloesch, Rishabh Joshi, Thomas Lampe, Junhyuk Oh, Nicolas
  Heess, Jonas Buchli, and Martin Riedmiller.
\newblock Learning from negative feedback, or positive feedback or both.
\newblock In \emph{International Conference on Learning Representations}, 2025.

\bibitem[Akbar et~al.(2021)Akbar, Robert, Pavlović, Jeliazkov, Snapkov,
  Slabodkin, Weber, Scheffer, Miho, Haff, Haug, Lund-Johansen, Safonova,
  Sandve, and Greiff]{Akbar2021-mk}
Rahmad Akbar, Philippe~A Robert, Milena Pavlović, Jeliazko~R Jeliazkov, Igor
  Snapkov, Andrei Slabodkin, Cédric~R Weber, Lonneke Scheffer, Enkelejda Miho,
  Ingrid~Hobæk Haff, Dag Trygve~Tryslew Haug, Fridtjof Lund-Johansen, Yana
  Safonova, Geir~K Sandve, and Victor Greiff.
\newblock A compact vocabulary of paratope-epitope interactions enables
  predictability of antibody-antigen binding.
\newblock \emph{Cell Rep.}, 34\penalty0 (11):\penalty0 108856, March 2021.

\bibitem[Balcan and Urner(2016)]{Balcan2016-lb}
Maria-Florina Balcan and Ruth Urner.
\newblock Active learning – modern learning theory.
\newblock In \emph{Encyclopedia of Algorithms}, pages 8--13. Springer New York,
  New York, NY, 2016.

\bibitem[Chaloner and Verdinelli(1995)]{Chaloner1995-hv}
Kathryn Chaloner and Isabella Verdinelli.
\newblock Bayesian experimental design: A review.
\newblock \emph{Stat. Sci.}, 10\penalty0 (3):\penalty0 273--304, 1995.

\bibitem[Elkan and Noto(2008)]{Elkan2008-ok}
Charles Elkan and Keith Noto.
\newblock Learning classifiers from only positive and unlabeled data.
\newblock In \emph{Proceedings of the 14th {ACM} {SIGKDD} international
  conference on Knowledge discovery and data mining}, New York, NY, USA, August
  2008. ACM.

\bibitem[Frazer et~al.(2021)Frazer, Notin, Dias, Gomez, Min, Brock, Gal, and
  Marks]{Frazer2021-uq}
Jonathan Frazer, Pascal Notin, Mafalda Dias, Aidan Gomez, Joseph~K Min, Kelly
  Brock, Yarin Gal, and Debora~S Marks.
\newblock Disease variant prediction with deep generative models of
  evolutionary data.
\newblock \emph{Nature}, 599\penalty0 (7883):\penalty0 91--95, 2021.

\bibitem[Gal et~al.(2017)Gal, Islam, and Ghahramani]{Gal2017-rp}
Yarin Gal, Riashat Islam, and Zoubin Ghahramani.
\newblock Deep bayesian active learning with image data.
\newblock In \emph{International conference on machine learning (ICML)}, 2017.

\bibitem[Genomics(2019)]{Genomics2019-ks}
10x Genomics.
\newblock A new way of exploring immunity - linking highly multiplexed antigen
  recognition to immune repertoire and phenotype, 2019.

\bibitem[Goda et~al.(2022)Goda, Hironaka, Kitade, and Foster]{Goda2022-pc}
Takashi Goda, Tomohiko Hironaka, Wataru Kitade, and Adam Foster.
\newblock Unbiased {MLMC} stochastic gradient-based optimization of bayesian
  experimental designs.
\newblock \emph{SIAM J. Sci. Stat. Comput.}, 44\penalty0 (1):\penalty0
  A286--A311, 2022.

\bibitem[Huan et~al.(2024)Huan, Jagalur, and Marzouk]{Huan2024-xa}
Xun Huan, Jayanth Jagalur, and Youssef Marzouk.
\newblock Optimal experimental design: Formulations and computations.
\newblock \emph{arXiv [stat.ME]}, July 2024.

\bibitem[Ipsen et~al.(2022)Ipsen, Mattei, and Frellsen]{Ipsen2022-cq}
Niels~Bruun Ipsen, Pierre-Alexandre Mattei, and Jes Frellsen.
\newblock How to deal with missing data in supervised deep learning?
\newblock In \emph{International Conference on Learning Representations
  (ICLR)}. orbit.dtu.dk, 2022.

\bibitem[Josse et~al.(2019)Josse, Chen, Prost, Scornet, and
  Varoquaux]{Josse2019-yq}
Julie Josse, Jacob~M Chen, Nicolas Prost, Erwan Scornet, and Gaël Varoquaux.
\newblock On the consistency of supervised learning with missing values.
\newblock \emph{arXiv [stat.ML]}, February 2019.

\bibitem[Kiryo et~al.(2017)Kiryo, Niu, Plessis, and Sugiyama]{Kiryo2017-ap}
Ryuichi Kiryo, Gang Niu, Marthinus C~du Plessis, and Masashi Sugiyama.
\newblock Positive-unlabeled learning with non-negative risk estimator.
\newblock March 2017.

\bibitem[Klebanoff et~al.(2023)Klebanoff, Chandran, Baker, Quezada, and
  Ribas]{Klebanoff2023-bx}
Christopher~A Klebanoff, Smita~S Chandran, Brian~M Baker, Sergio~A Quezada, and
  Antoni Ribas.
\newblock {T} cell receptor therapeutics: immunological targeting of the
  intracellular cancer proteome.
\newblock \emph{Nat. Rev. Drug Discov.}, pages 1--22, October 2023.

\bibitem[Lindley(1956)]{Lindley1956-ig}
D~V Lindley.
\newblock On a measure of the information provided by an experiment.
\newblock \emph{Ann. Math. Stat.}, 27\penalty0 (4):\penalty0 986--1005,
  December 1956.

\bibitem[Miller(2021)]{Miller2021-xj}
Jeffrey~W Miller.
\newblock Asymptotic normality, concentration, and coverage of generalized
  posteriors.
\newblock \emph{J. Mach. Learn. Res.}, 22\penalty0 (168):\penalty0 1--53, 2021.

\bibitem[Pavlović et~al.(2021)Pavlović, Scheffer, Motwani, Kanduri, Kompova,
  Vazov, Waagan, Bernal, Costa, Corrie, Akbar, Al~Hajj, Balaban, Brusko,
  Chernigovskaya, Christley, Cowell, Frank, Grytten, Gundersen, Haff, Hovig,
  Hsieh, Klambauer, Kuijjer, Lund-Andersen, Martini, Minotto, Pensar, Rand,
  Riccardi, Robert, Rocha, Slabodkin, Snapkov, Sollid, Titov, Weber, Widrich,
  Yaari, Greiff, and Sandve]{Pavlovic2021-sn}
Milena Pavlović, Lonneke Scheffer, Keshav Motwani, Chakravarthi Kanduri,
  Radmila Kompova, Nikolay Vazov, Knut Waagan, Fabian L~M Bernal,
  Alexandre~Almeida Costa, Brian Corrie, Rahmad Akbar, Ghadi~S Al~Hajj, Gabriel
  Balaban, Todd~M Brusko, Maria Chernigovskaya, Scott Christley, Lindsay~G
  Cowell, Robert Frank, Ivar Grytten, Sveinung Gundersen, Ingrid~Hobæk Haff,
  Eivind Hovig, Ping-Han Hsieh, Günter Klambauer, Marieke~L Kuijjer, Christin
  Lund-Andersen, Antonio Martini, Thomas Minotto, Johan Pensar, Knut Rand,
  Enrico Riccardi, Philippe~A Robert, Artur Rocha, Andrei Slabodkin, Igor
  Snapkov, Ludvig~M Sollid, Dmytro Titov, Cédric~R Weber, Michael Widrich, Gur
  Yaari, Victor Greiff, and Geir~Kjetil Sandve.
\newblock The {immuneML} ecosystem for machine learning analysis of adaptive
  immune receptor repertoires.
\newblock \emph{Nature Machine Intelligence}, 3\penalty0 (11):\penalty0
  936--944, November 2021.

\bibitem[Plessis et~al.(2015)Plessis, Niu, and Sugiyama]{Plessis2015-ks}
Marthinus~Du Plessis, Gang Niu, and Masashi Sugiyama.
\newblock Convex formulation for learning from positive and unlabeled data.
\newblock In Francis Bach and David Blei, editors, \emph{Proceedings of the
  32nd International Conference on Machine Learning}, volume~37 of
  \emph{Proceedings of Machine Learning Research}, pages 1386--1394, Lille,
  France, 2015. PMLR.

\bibitem[Rainforth et~al.(2018)Rainforth, Cornish, Yang, Warrington, and
  Wood]{Rainforth2018-qp}
Tom Rainforth, Robert Cornish, Hongseok Yang, Andrew Warrington, and Frank
  Wood.
\newblock On nesting monte carlo estimators.
\newblock In \emph{International Conference on Machine Learning (ICML)}, 2018.

\bibitem[Rainforth et~al.(2024)Rainforth, Foster, Ivanova, and
  Smith]{Rainforth2024-cn}
Tom Rainforth, Adam Foster, Desi~R Ivanova, and Freddie~Bickford Smith.
\newblock Modern bayesian experimental design.
\newblock \emph{Stat. Sci.}, 2024.

\bibitem[Skora et~al.(2015)Skora, Douglass, Hwang, Tam, Blosser, Gabelli, Cao,
  Diaz, Papadopoulos, Kinzler, Vogelstein, and Zhou]{Skora2015-qo}
Andrew~D Skora, Jacqueline Douglass, Michael~S Hwang, Ada~J Tam, Richard~L
  Blosser, Sandra~B Gabelli, Jianhong Cao, Luis~A Diaz, Jr, Nickolas
  Papadopoulos, Kenneth~W Kinzler, Bert Vogelstein, and Shibin Zhou.
\newblock Generation of {MANAbodies} specific to {HLA}-restricted epitopes
  encoded by somatically mutated genes.
\newblock \emph{Proc. Natl. Acad. Sci. U. S. A.}, 112\penalty0 (32):\penalty0
  9967--9972, August 2015.

\bibitem[Smith et~al.(2023)Smith, Kirsch, Farquhar, Gal, Foster, and
  Rainforth]{Smith2023-dt}
Freddie~Bickford Smith, Andreas Kirsch, Sebastian Farquhar, Yarin Gal, Adam
  Foster, and Tom Rainforth.
\newblock Prediction-oriented bayesian active learning.
\newblock \emph{arXiv [cs.LG]}, April 2023.

\bibitem[Song et~al.(2021)Song, Bremer, Hinds, Raskutti, and
  Romero]{Song2021-rx}
Hyebin Song, Bennett~J Bremer, Emily~C Hinds, Garvesh Raskutti, and Philip~A
  Romero.
\newblock Inferring protein sequence-function relationships with large-scale
  positive-unlabeled learning.
\newblock \emph{Cell Syst.}, 12\penalty0 (1):\penalty0 92--101.e8, January
  2021.

\bibitem[Thadani et~al.(2023)Thadani, Gurev, Notin, Youssef, Rollins, Ritter,
  Sander, Gal, and Marks]{Thadani2023-hc}
Nicole~N Thadani, Sarah Gurev, Pascal Notin, Noor Youssef, Nathan~J Rollins,
  Daniel Ritter, Chris Sander, Yarin Gal, and Debora~S Marks.
\newblock Learning from prepandemic data to forecast viral escape.
\newblock \emph{Nature}, 622\penalty0 (7984):\penalty0 818--825, October 2023.

\bibitem[Vaicenavicius et~al.(2019)Vaicenavicius, Widmann, Andersson, Lindsten,
  Roll, and Sch{\"o}n]{Vaicenavicius2019-ir}
Juozas Vaicenavicius, David Widmann, Carl Andersson, Fredrik Lindsten, Jacob
  Roll, and Thomas~B Sch{\"o}n.
\newblock Evaluating model calibration in classification.
\newblock February 2019.

\bibitem[Watson et~al.(2023)Watson, Juergens, Bennett, Trippe, Yim, Eisenach,
  Ahern, Borst, Ragotte, Milles, Wicky, Hanikel, Pellock, Courbet, Sheffler,
  Wang, Venkatesh, Sappington, Torres, Lauko, De~Bortoli, Mathieu, Ovchinnikov,
  Barzilay, Jaakkola, DiMaio, Baek, and Baker]{Watson2023-sp}
Joseph~L Watson, David Juergens, Nathaniel~R Bennett, Brian~L Trippe, Jason
  Yim, Helen~E Eisenach, Woody Ahern, Andrew~J Borst, Robert~J Ragotte, Lukas~F
  Milles, Basile I~M Wicky, Nikita Hanikel, Samuel~J Pellock, Alexis Courbet,
  William Sheffler, Jue Wang, Preetham Venkatesh, Isaac Sappington,
  Susana~Vázquez Torres, Anna Lauko, Valentin De~Bortoli, Emile Mathieu,
  Sergey Ovchinnikov, Regina Barzilay, Tommi~S Jaakkola, Frank DiMaio, Minkyung
  Baek, and David Baker.
\newblock De novo design of protein structure and function with {RFdiffusion}.
\newblock \emph{Nature}, 620\penalty0 (7976):\penalty0 1089--1100, August 2023.

\bibitem[Weinstein et~al.(2022)Weinstein, Amin, Grathwohl, Kassler, Disset, and
  Marks]{Weinstein2022-sw}
Eli~N Weinstein, Alan~N Amin, Will Grathwohl, Daniel Kassler, Jean Disset, and
  Debora~S Marks.
\newblock Optimal design of stochastic {DNA} synthesis protocols based on
  generative sequence models.
\newblock In \emph{Proceedings of the 25th International Conference on
  Artificial Intelligence and Statistics ({AISTATS})}. PMLR, 2022.

\bibitem[Weinstein et~al.(2024)Weinstein, Gollub, Slabodkin, Gardner, Dobbs,
  Cui, Amin, Church, and Wood]{Weinstein2024-wm}
Eli~N Weinstein, Mattia~G Gollub, Andrei Slabodkin, Cameron~L Gardner, Kerry
  Dobbs, Xiao-Bing Cui, Alan~N Amin, George~M Church, and Elizabeth~B Wood.
\newblock Manufacturing-aware generative model architectures enable biological
  sequence design and synthesis at petascale.
\newblock \emph{bioRxiv}, page 2024.09.13.612900, September 2024.

\bibitem[Wu(2021)]{Wu2021-fx}
Yifan Wu.
\newblock \emph{Learning to Predict and Make Decisions under Distribution
  Shift}.
\newblock PhD thesis, Carnegie Mellon University, 2021.

\bibitem[Zemplenyi and Miller(2023)]{Zemplenyi2023-cd}
Michele Zemplenyi and Jeffrey~W Miller.
\newblock Bayesian optimal experimental design for inferring causal structure.
\newblock \emph{Bayesian Anal.}, 18\penalty0 (3):\penalty0 929--956, 2023.

\end{thebibliography}

\clearpage
\appendix
\thispagestyle{empty}

\onecolumn
\aistatstitle{Accelerated Learning on Large Scale Screens: \\
Supplementary Materials}

\section{Evaluation Details}

\subsection{Accuracy Approximation} \label{apx:accuracy}

To approximate the accuracy $\textsc{A}_\pr[t_\theta]$, let $(\tilde x_1, 1), ..., (\tilde x_{\tilde n}, 1), y_{\tilde n +1}, ..., y_{\tilde N}$ denote heldout data, and draw samples $x'_1, ..., x'_M \sim \pr(x)$. We obtain,
\begin{align}
\textsc{A}_\pr[t_\theta] =& \pr(y=1) \mathbb{E}[t_\theta(X)|Y=1] + \mathbb{E}_\pr[1 - t_\theta(X)] - \pr(y=1) \mathbb{E}_p[1 - t_\theta(X)|Y=1]\\
\approx &(\frac{1}{\tilde N}\sum_{i=1}^{\tilde N} \tilde y_i)\frac{1}{\tilde n} \sum_{i=1}^{\tilde n} t_\theta(\tilde x_i) + \frac{1}{M} \sum_{j=1}^M (1 - t_\theta(x'_i)) - (\frac{1}{\tilde N}\sum_{i=1}^{\tilde N} \tilde y_i)\frac{1}{\tilde n} \sum_{i=1}^{\tilde n} (1 - t_\theta(\tilde x_i))
\end{align}

\paragraph{Precision and Recall} \label{apx:precision-recall}

We can apply the same strategy as used for accuracy to approximate the precision and recall, along with other standard measures of classifier performance.
The precision admits the decomposition
\begin{equation}
    \frac{\Eb_\pr[Y t_\theta(X)]}{\Eb_\pr[t_\theta(X)]} = \frac{\pr(y=1) \mathbb{E}[t_\theta(X)|Y=1]}{\Eb_\pr[t_\theta(X)]}
\end{equation}
The numerator can be estimated as for accuracy; the denominator can be estimated using samples from $\pr(x)$. The recall is the same, but with a different denominator,
\begin{equation}
    \frac{\Eb_\pr[Y t_\theta(X)]}{\pr(y=1)}.
\end{equation}
We can estimate the denominator using y-only data, as for accuracy.

\subsection{Calibration Approximation} \label{apx:calibration}

We form an estimate of the histogrammed ECE,
\begin{equation}
	\textsc{ece}_\pr[\pr_{\theta}] \approx \sum_{k=1}^K \pr(\pr_{\theta}(y=1|X) \in \Phi_k) d\left( \mathbb{E}_{\pr}[Y \mid \pr_{\theta}(y=1|X) \in \Phi_k], \mathbb{E}_{\pr}[\pr_{\theta}(y=1|X) \mid \pr_{\theta}(y=1|X) \in \Phi_k]\right)
\end{equation}
Let $(\tilde x_1, 1), ..., (\tilde x_{\tilde n}, 1), y_{\tilde n +1}, ..., y_{\tilde N}$ denote heldout data, and draw samples $x'_1, ..., x'_M \sim \pr(x)$. We obtain,
\begin{align}
	\pr(\pr_{\theta}(y=1\mid X) \in \Phi_k) &\approx \frac{1}{M} \sum_{j=1}^M \mathbb{I}(\pr_{\theta}(y=1\mid x'_j) \in \Phi_k)\\
	\mathbb{E}_{\pr}[Y \mid \pr_{\theta}(y=1\mid X) \in \Phi_k] &= \frac{\mathbb{E}_{\pr}[Y \mathbb{I}(\pr_\theta(y=1\mid X) \in \Phi_k)]}{\mathbb{E}_{\pr}[\mathbb{I}(\pr_\theta(y=1\mid X) \in \Phi_k)]}\\
	&\approx \frac{\left(\frac{1}{\tilde N} \sum_{i=1}^{\tilde N} \tilde y_i\right) \frac{1}{\tilde n} \sum_{i= 1}^{\tilde n} \mathbb{I}(\pr_\theta(y=1\mid \tilde x_i) \in \Phi_k)}{\frac{1}{M} \sum_{j=1}^M \mathbb{I}(\pr_\theta(y=1\mid x'_j) \in \Phi_k)}\\
	\mathbb{E}_{\pr}[\pr_{\theta}(y=1\mid X) \mid \pr_{\theta}(y=1\mid X) \in \Phi_k] & =  \frac{\mathbb{E}_{\pr}[\pr_{\theta}(y=1\mid X) \mathbb{I}(\pr_{\theta}(y=1\mid X) \in \Phi_k)]}{\mathbb{E}_{\pr}[\mathbb{I}(\pr_{\theta}(y=1\mid X) \in \Phi_k)]}\\
	&\approx \frac{\frac{1}{M} \sum_{j=1}^M \pr_\theta(y=1\mid x'_j)\mathbb{I}(\pr_\theta(y=1\mid x'_i) \in \Phi_k)}{\frac{1}{M} \sum_{j=1}^M \mathbb{I}(\pr_\theta(y=1\mid x'_j) \in \Phi_k)}
\end{align}

\section{Proofs} 

\subsection{Proof of \Cref{thm:consistent} and \Cref{thm:BvM}} \label{apx:consistent}

We assume the distribution $\pr(x)$ has support over all possible values of $x$, i.e. we will only be able to learn $\pr(y \mid x)$ for $x$ that can actually occur.
\begin{assumption}[Full support] \label{asm:support}
    $\pr(x) > 0$ for all $x \in \Xc$.
\end{assumption}

Our results build on the observation that we can nonparametrically identify $\pr(y \mid x)$ from $\pr(x)$ (the library distribution), $\pr(x \mid y=1)$ (the distribution of positive examples) and $\pr(y=1)$ (the activity/hit rate).
\begin{lemma}[Nonparametric identification] \label{thm:nonparam-id}
    Let \Cref{asm:support} hold. Given $\pr(x)$, $\pr(x \mid y=1)$ and $\pr(y=1)$, we can compute $\pr(y \mid x)$.
\end{lemma}
\begin{proof}
    \begin{equation} \label{eqn:bayes_rule}
        \pr(y =1 \mid x) = \frac{\pr(x \mid y=1) \pr(y=1)}{\pr(x)}
    \end{equation}
\end{proof}

To prove \Cref{thm:consistent} and \Cref{thm:BvM}, we make use of the average log likelihood of the model,
\begin{equation}
	L_{q,n}(\theta) \triangleq \frac{1}{n} \sum_{i=1}^{nq} \log \pr_\theta(y = 1 \mid x_i) + \frac{1}{n} \sum_{i=nq + 1}^{n} \log \pr_\theta(y = 0 \mid x_i).
\end{equation}
By the strong law of large numbers, $L_{q,n}(\theta)$ converges pointwise a.s. to the expected log likelihood as $n \to \infty$,
\begin{align}
	L_q(\theta) &\triangleq q \Eb_{\pr(x \mid y=1)}[\log \pr_\theta(y=1\mid X)] + (1-q)\Eb_{\pr(x \mid y=0)}[\log \pr_\theta(y=0\mid X)]\\
	&= \Eb_{\pr(x \mid y)\pr_q(y) }[\log \pr_\theta(Y \mid X)],
\end{align}
where $\pr_q(y)$ is defined as the Bernoulli distribution with parameter $q$.

Our key intermediate result is that the expected log likelihood of the model under the data distribution is maximized at $\theta_0$, for any $q$.
\begin{proposition}[$\theta_0$ is optimal] \label{thm:theta0_id}
    Let \Cref{asm:specified}, \Cref{asm:known_hit} and \Cref{asm:support} hold. Then for all $q \in [0, 1]$, $L_q(\theta)$ is maximized at $\theta_0$.
\end{proposition}
\begin{proof}
Let $\pr_\theta(x \mid y) \propto \pr_\theta(y \mid x)\pr(x)$ denote the conditional distribution of $y$ under the model and the true $x$ distribution. We have
\begin{align}
	\underset{\theta \in \Theta}{\argmax}\, L_q(\theta) &= \underset{\theta \in \Theta}{\argmax}\, \Eb_{\pr(x \mid y)\pr_q(y) }[\log \pr_\theta(Y\mid  X) + \log \pr(X)]\\
	&= \underset{\theta \in \Theta}{\argmax}\, \Eb_{\pr(x \mid y)\pr_q(y) }[\log \pr_\theta(X\mid  Y) + \log \pr(Y)]\\
&= \underset{\theta \in \Theta}{\argmax}\, \Eb_{\pr(x \mid y) \pr_q(y)}[\log \pr_\theta(X\mid  Y)] \\
    &= \underset{\theta \in \Theta}{\argmin}\, q \mathrm{KL}(\pr(x \mid y=1) \| \pr_\theta(x \mid y=1)) + (1-q) \mathrm{KL}(\pr(x \mid y=0) \| \pr_\theta(x \mid y=0)) \label{eqn:kl_minimizer}
\end{align} 
where in the first line we use the fact that $\pr(x)$ does not depend on $\theta$, for the second line we use the chain rule for conditional entropy, for the third line we use \Cref{asm:known_hit}.

When $q > 0$, the minimum of the first term in \Cref{eqn:kl_minimizer} is reached when $\pr_\theta(x \mid y=1) = \pr(x \mid y=1)$. 
This occurs if and only if $\pr_\theta(y \mid x) = \pr(y \mid x)$, by \Cref{eqn:bayes_rule}, since $\pr(x) > 0$ for all $x$ and $\pr_\theta(y) = \pr(y)$ (\Cref{asm:known_hit}).
So, since the model is well specified (\Cref{asm:specified}) $\theta_0$ must be a minimizer of the first term in \Cref{eqn:kl_minimizer}.
The same argument holds for the second term of \Cref{eqn:kl_minimizer}, so $\theta_0$  must minimize \Cref{eqn:kl_minimizer} overall. 
When $q = 0$, the same argument holds with $y=0$ switched for $y=1$.
\end{proof}

For this result to translate into posterior consistency and asymptotic normality, we must place regularity conditions on the model's parameterization. 
We use the conditions of Theorem 3.2 in \cite{Miller2021-xj}, which provides a version of the Bernstein-von Mises theorem (BvM) which handles settings where the model's distribution does not match the data distribution, as is the case here. \begin{assumption}[Bernstein-von Mises assumptions] \label{asm:BvM}
Let $\Theta \subseteq \Rb^d$. Assume the prior $\pr(\theta)$ is continuous at $\theta_0$ and $\pi(\theta_0) > 0$. 
Let $E \subseteq \Theta$ be a convex bounded open set that contains $\theta_0$.
Assume $L_{n, q}(\theta)$ has continuous third derivatives on $E$. Assume $-\nabla_\theta^2 L_q(\theta_0)$ is positive definite.
Assume the set $\{\big(\sum_{i,j,k=1}^d \left(\frac{\partial L_{n,q}}{\partial \theta_i \partial \theta_j \partial \theta_k}(\theta)\right)^2\big)^{1/2}: \theta \in E, n \in \mathbb{N} \}$ is bounded.
Assume that either (a) $L_q(\theta) < L_q(\theta_0)$ for all $\theta \in K \setminus \{\theta_0\}$ and $\limsup_n \sup_{\theta \in \Theta \setminus K} L_{n,q}(\theta) < L_{q}(\theta_0)$ for a compact $K \subseteq E$ with $\theta_0$ in the interior of $K$ or 
	(b) $-L_{n,q}(\theta)$ is convex for each $n$ and $\nabla L(\theta_0) = 0$.
\end{assumption}

Crucially, the reason \Cref{asm:BvM} can be satisfied here is because of \Cref{thm:theta0_id}, that is, $\theta_0$ is a maximizer of the expected log likelihood, and hence we have $\nabla L(\theta_0) = 0$ and $-\nabla_\theta^2 L_q(\theta_0)$ positive definite.
With \Cref{asm:BvM} in place, \Cref{thm:consistent} and \Cref{thm:BvM} follow from Theorem 3.2 in \cite{Miller2021-xj}.

\subsection{Proof of \Cref{thm:inconsistent}} \label{apx:inconsistent}

If we do not have the constraint of \Cref{asm:known_hit}, the learned model will reflect the shifted distribution $\pr(x \mid y) \pr_q(y)$ rather than the target distribution $\pr(x, y)$. 
We assume first that the model includes this shifted distribution. Let $\pr_q(x) = \sum_{y=0}^1 \pr(x \mid y) \pr_q(y)$ and $\pr_q(y \mid x) = \pr(x \mid y)\pr_q(y)/\pr_q(x)$ denote the marginal and conditional under the shifted distribution.
\begin{assumption}[Specified under shifts] \label{asm:shift-spec}
    Assume there exists a $\tilde \theta \in \Theta$ such that $\pr_{\tilde \theta}(y \mid x) = \pr_q(y \mid x) $ for all $x,y$.
\end{assumption}
For this assumption to hold at $q \neq \pr(y=1)$, \Cref{asm:known_hit} must be violated, since $\int \pr_q(y=1 \mid x) \pr(x) dx = \pr_q(y=1) = q$. This means, moreover, that $\pr_{\tilde \theta}(y \mid x) \neq \pr(y \mid x)$. On the other hand, when $q= \pr(y=1)$, \Cref{asm:shift-spec} coincides with \Cref{asm:specified}, and we can set $\tilde \theta = \theta_0$.

Now, the expected likelihood is maximized at $\pr_q(y \mid x)$.
\begin{proposition}[$\tilde \theta$ is optimal] \label{thm:tilde_theta_id}
    Let \Cref{asm:shift-spec} and \Cref{asm:support} hold. Then $L_q(\theta)$ is maximized at $\tilde \theta$.
\end{proposition}
\begin{proof}
     We have,
    \begin{align}
        \underset{\theta \in \Theta}{\argmax}\, L_q(\theta) &= \underset{\theta \in \Theta}{\argmax}\, \Eb_{\pr(x \mid y)\pr_q(y) }[\log \pr_\theta(Y\mid  X)]\\
        &= \underset{\theta \in \Theta}{\argmin} \, \Eb_{\pr_q(x)}[\mathrm{KL}( \pr_q(Y \mid X) \| \pr_\theta(Y \mid X))]
    \end{align}
    The KL divergence is minimized whenever $\pr_q(Y \mid X) = \pr_\theta(Y \mid X)$.
\end{proof}

Now, assume the model is sufficiently regular around $\tilde \theta$.
\begin{assumption}[BvM assumptions] \label{asm:bvm-inconsistent}
    \Cref{asm:BvM} holds with $\theta_0$ replaced by  $\tilde \theta$.
\end{assumption}

As before, this assumption can only be satisfied because of \Cref{thm:tilde_theta_id}. Then, \Cref{thm:inconsistent} follows from Theorem 3.2 in \cite{Miller2021-xj}.

\subsection{Proof of \Cref{thm:argmax_1}} \label{apx:argmax_1}
\begin{proof}
	Using the formula for the entropy of a Gaussian, we can see
	\begin{equation}
	\underset{q \in [0, 1]}{\argmin}\, \mathcal{H}(\mathcal{N}(0, H_q^{-1})) = \underset{q \in [0, 1]}{\argmax} \, \det H_q
\end{equation}
We now have 
\begin{align}
	H_q = & -q\Eb_{\pr(x \mid y=1)}[\nabla^2_\theta \log \pr_\theta(y=1 \mid X) |_{\theta_0}] - (1-q) \Eb_{\pr(x \mid y=0)}[\nabla^2_\theta \log \pr_\theta(y=0 \mid X) |_{\theta_0}]\\
=& q \Ic_1 + (1-q) \pr(S_0 \mid y=0) \Ic_0 \label{eqn:Hq}
\end{align}
where in the second line we have used that $\pr(x \mid y = 1) = \pr(x \mid S_0, y=1)$  by \Cref{eqn:sparse_active}. This implies $H_1 = \Ic_1$ so we have,
\begin{equation}
	H_q = H_1 + (1-q) (\pr(S_0 \mid y=0) \Ic_0 - \Ic_1).
\end{equation}
Recall some properties of the determinant: $\det (A + B) \ge \det A + \det B$, and $\det a B = a^d \det B$ for scalar $a$. We obtain,
\begin{align}
	\det H_1 &\ge \det H_q + (1-q)^d \det (\Ic_1 - \pr(S_0 \mid y=0) \Ic_0)\\
	&\ge \det H_q + (1-q)^d \det (\Ic_1 - \frac{\eta}{1 - \eta} \Ic_0) + (1-q)^d\left(\frac{\eta}{1-\eta}  - \pr(S_0 \mid y=0)\right)^d\det \Ic_0
\end{align}
Now we have, 
\begin{align}
	\pr(S_0 \mid y=0)= \frac{\pr(S_0, y=0)}{\pr(y=0)} = \frac{\eta \pr(y=0 \mid S_0)}{1-\eta + \eta \pr(y=0 \mid S_0)} \le \frac{\eta}{1 - \eta}. \label{eqn:eta_bound}
\end{align}
where we have used \Cref{eqn:sparse_active} for the second equality, and $0 \le \pr(y=0 \mid S_0) \le 1$ for the inequality. 
Since $\Ic_0$ is positive definite by assumption, we have
\begin{equation}
	\det H_1 \ge \det H_q + (1-q)^d \det (\Ic_1 - \frac{\eta}{1 - \eta} \Ic_0)
\end{equation}
Applying the assumption on the second term, we find that for any $0 \le q < 1$,
\begin{equation}
	\det H_1 > \det H_q.
\end{equation}

Note $\Ic_0$ and $\Ic_1$ do not depend on $\eta$, as they condition on $S_0$.
So, since $\Ic_1$ and $\Ic_0$ are positive definite, there always exists $\eta$ sufficiently small such that $\det(\Ic_1 - \frac{\eta}{1 - \eta} \Ic_0) > 0$. 
\end{proof}

An interesting feature of this result is the optimal choice of $q$ exhibits a threshold effect as a function of $\eta$. 
Naively, one may hypothesize that the optimal $q$ converges to $1$ as $\eta$ converges to 0. 
But the result here is stronger: for sufficiently small $\eta$, the optimal choice of $q$ becomes \emph{exactly} 1.
This means that if we expect the optimal region to be small, we should collect \textit{only} positive examples.

\subsection{Back-of-the-envelope information gain} \label{apx:info_gain}

We can compute the information gain as the difference in Shannon entropy of the asymptotic posterior distribution, when setting $q=1$ versus $q=\pr(y=1)$ (or, equivalently, drawing samples i.i.d. from $\pr(x, y)$):
\begin{equation}
    \mathcal{H}(\mathcal{N}(0, n^{-1} H_q^{-1})) - \mathcal{H}(\mathcal{N}(0, n^{-1} H_1^{-1}))
\end{equation}

Plugging in \Cref{eqn:Hq}, and using $\Ic \triangleq \Ic_0 = \Ic_1$ under our back-of-the-envelope assumptions, we have
\begin{align}
    =\frac{1}{2} \log \det \Ic - \frac{1}{2} \log \det [q \Ic + (1-q) \pr(S_0 \mid y=0) \Ic]\\
    = - \frac{d}{2} \log(q + (1-q) \pr(S \mid y=0))
\end{align}
Applying \Cref{eqn:eta_bound}, the information gain is at least
\begin{align}
    \ge - \frac{d}{2} \log (q + (1-q) \frac{\eta}{1-\eta})
\end{align}
Plugging in the back-of-the-envelope assumption $q = \pr(y=1) = \eta/2$, we have
\begin{align}
    &= - \frac{d}{2} \log q - \frac{d}{2} \log(1 + 2 \frac{1-q}{1 - 2q})\\
    &\ge - \frac{d}{2} \log 3q.
\end{align}

We can compare this information gain from increasing the sample size from $n$ to $c n$,
\begin{align}
    \mathcal{H}(\mathcal{N}(0, n^{-1} H_q^{-1})) - \mathcal{H}(\mathcal{N}(0, (c n)^{-1} H_q^{-1})) =& \frac{1}{2} \log [(c n)^d] - \frac{1}{2} \log [n^d]\\
    &= \frac{d}{2} \log c
\end{align}
So, back-of-the envelope, collecting positive examples provides at least as much information as increasing the sample size from $n$ to $\frac{n}{3 \pr(y=1)}$. 

\section{Empirical Results}

\subsection{Models and training} \label{apx:deep-models}
In all three datasets, we used a shallow single-layer CNN with a kernel size of 5, stride 1, 32 channels, and a 16-dimensional initial embedding for the individual amino acids. Where possible (in the synthetic and the large-scale demonstration on antibodies) we used a batch size of 128 positive examples; for the semi-experimental dataset, due to low $n$, we used a batch size of 16 positive examples. The models were trained for 1000 epochs on a single NVIDIA H100 GPU (about 6 minutes per dataset). For the large-scale demonstration on antibodies, we additionally trained a transformer-based model: encoder-only, depth 8, attention dimension 16, 16 eight-dimensional attention heads.

\begin{figure*}[t]
    \centering
    \includegraphics[width=0.3\linewidth]{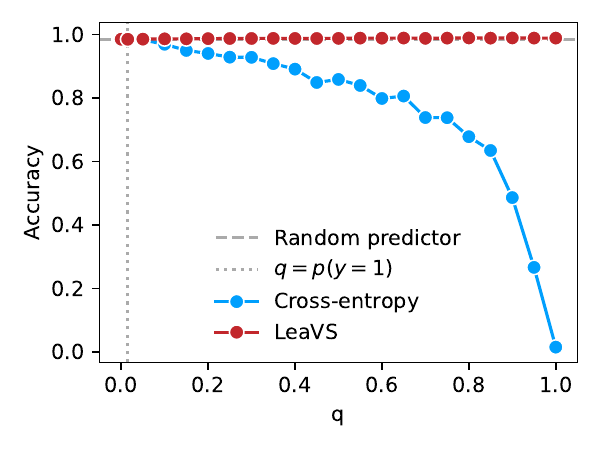}
    \caption{Accuracy on synthetic data (\Cref{fig:accuracy_zoomed}) with y-axis extended to zero.} \label{fig:accuracy_full}
\end{figure*}

\begin{figure*}[t]
    \centering
    \begin{subfigure}[t]{0.4\linewidth}
        \includegraphics[width=\linewidth]{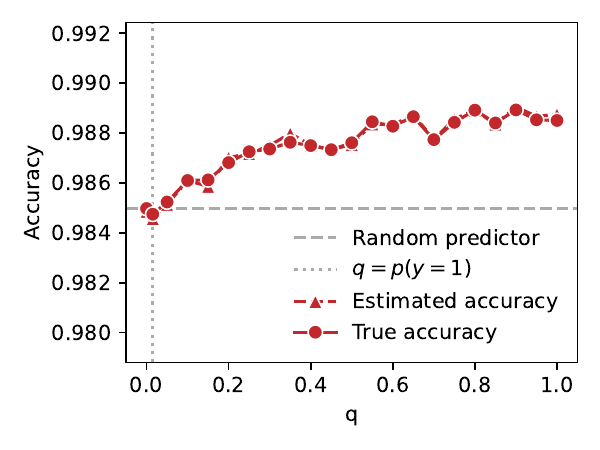}
    \caption{Accuracy.} \label{fig:accuracy_estimated}
    \end{subfigure}
    \begin{subfigure}[t]{0.4\linewidth}
    \centering
    \includegraphics[width=\linewidth]{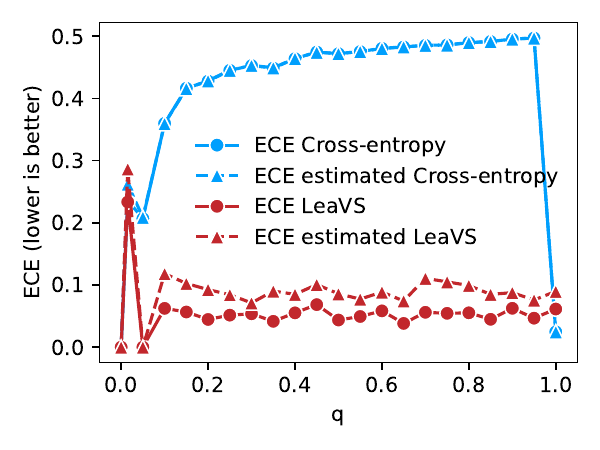}
    \caption{Expected calibration error.} \label{fig:calibration}
    \end{subfigure}
    \caption{Evaluation estimates using heldout positives-only data, on the synthetic data.} \end{figure*}

\subsection{Synthetic data} \label{apx:synthetic}
In the synthetic data setting, we set $\pr(x)$ to be a variational synthesis model of antibody CDRH3 amino acid sequences \citep{Weinstein2024-wm}. We set $\pr(y \mid x)$ so that $y$ depends on the presence of specific amino acid sequences at specific positions. Namely, $\pr(y \mid x)$ is defined as follows: given a sequence $x$, we set its 'base strength' $s(x)$ to 1. Then, we check for the presence of the following substrings: (i) \texttt{"P"} or \texttt{"C"} at position 3 (zero-indexed), (ii) \texttt{"N"} or \texttt{"C"} at position 5, and (iii) "\texttt{"PC"} or \texttt{"SS"} starting at position 6. The presence of each substring increases the strength $s(x)$ by a multiplier of 10. We then add noise by sampling the final value $c(x)$ value from the Negative Binomial distribution with a mean of $s(x)$ and variance $s(x) + \frac{s(x)^2}{\phi}$ (where $\phi$ = 2.28), and set $y = \mathbb{I}[c(x) > 30]$ --- so that a sequence is required to have at least two of the patterns for its label to be positive (but some of the sequences that only have a single pattern may be labeled positive as well due to added stochasticity). In this setting $\pr(y = 1) = 0.015$. This process, including the choice of $\phi$, was designed based on the experimental conditions in the antibody demonstration. We used $n = 2000$ observed sequences for training, with $q$ ranging from $0$ to $1$ with a step of $0.05$, with an additional $q = \pr(y = 1)$. For testing, we used a set of 400\,000 sequences sampled from $\pr(x)$. For evaluating the estimated accuracy and calibration, we split this set into halves, and used only the positive examples from one half to represent our observed set, while the other half represented the unlabeled set (the library). We used 10 equal-sized bins for evaluation of the the estimated calibration error.

\begin{figure*}[t]
    \centering
    \begin{subfigure}[t]{0.4\linewidth}
        \includegraphics[width=\linewidth]{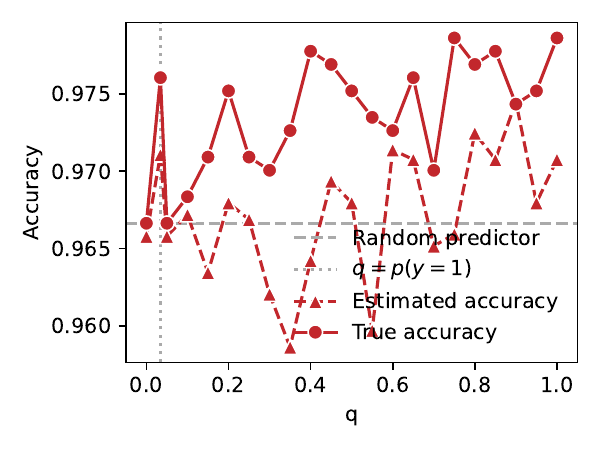}
    \caption{Accuracy.} 
    \end{subfigure}
    \begin{subfigure}[t]{0.4\linewidth}
    \centering
    \includegraphics[width=\linewidth]{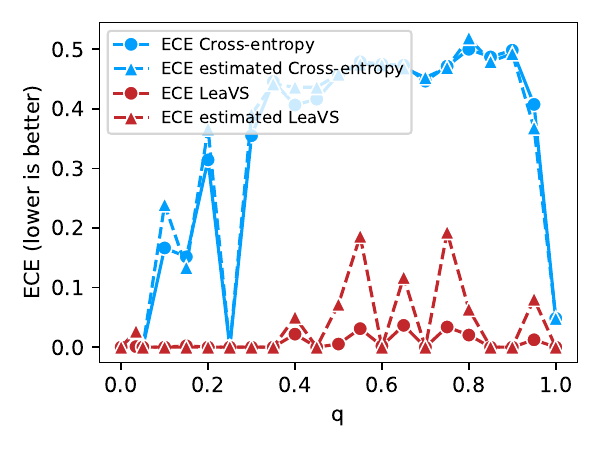}
    \caption{Expected calibration error.} 
    \end{subfigure}
    \caption{Evaluation estimates using heldout positives-only data, on the experimentally measured TCR data.} \label{fig:semiexperiment-evals}
\end{figure*}

\begin{figure*}[t]
    \centering
    \includegraphics[width=0.3\linewidth]{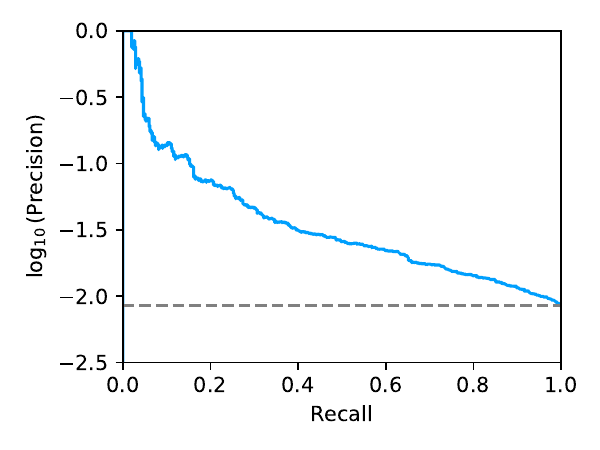}
    
    \caption{Predicting binding of TCR mimicking scFv CAR therapeutic candidates  to MAGE-A4, a challenging oncology target. Here we plot the precision recall curve using sequenced samples from the variational synthesis library as samples from $\pr(x)$, rather than samples from the computational library model as in \Cref{fig:pr_curve}. The AUC is 0.07.}
    \label{fig:transformer_experiment_sequenced}
\end{figure*}

\subsection{Experimental TCR data} \label{apx:semi-experimental}
For the experimental data setting, we used a dataset of T cell receptors (TCRs) from human CD8+ T cells, screened for binding against multiple targets, publicly available at \url{https://www.10xgenomics.com/datasets/cd-8-plus-t-cells-of-healthy-donor-1-1-standard-3-0-2} \citep{Genomics2019-ks}. We focused on a single target – the influenza antigen. In this dataset, we have measurements of $(x, y)$ for each example but we do not have control over $\pr(x)$, as the sequences come from a human patient. Instead, we split the data set into halves and used the empirical distribution of one half as an approximation of $\pr(x)$ (note this strategy would also be feasible in practice, since sequencing TCRs is substantially easier than obtaining TCR-binding activity measurements). We used the other half of the complete dataset to generate smaller datasets with varying values of $q$. Note that because we must subsample a full dataset and $p(y) = 0.034$, $n$ is relatively small, which increases variability. In total, there were 400 positive cells, 10\% (40 positive cells) of the full dataset is reserved for the heldout set, which leaves 360 cells. The full training set was split in halves (without stratifying by label) as described above --- which left us with $n = 170$ positive cells in the training set.

\subsection{Large scale demonstration} \label{apx:vs-demo}

For the large-scale demonstration, we used an experimental dataset of 9000 observed antibody CDRH3 sequences, paired with integer-valued binding measurements against 9 antigens – these sequences are the outcome of a high-throughput screen of a variational synthesis library of antibody CDRH3 amino acid sequences \citep{Weinstein2024-wm}. 
(Further details of the wetlab experiment will appear in a forthcoming publication.)
Each observed sequences is positive against at least one of the targets. We used 10\% of the data for the heldout set, and we used the corresponding variational synthesis model as $\pr(x)$. 
We use a negative binomial likelihood in $\pr_\theta(y \mid x)$
To compute precision and recall for a non-binary outcome $Y$, we use a sliding threshold: we replace $Y$ with $\mathbb{I}(Y > \tau)$ in \Cref{apx:accuracy}, where $\tau$ is the same threshold as for the predictor, $t_\theta(X) = \mathbb{I}(\pr_\theta(y=1 \mid x) > \tau)$.

\begin{figure*}[t]
    \centering
    \begin{subfigure}[t]{0.3\linewidth}
    \centering
    \includegraphics[width=\linewidth]{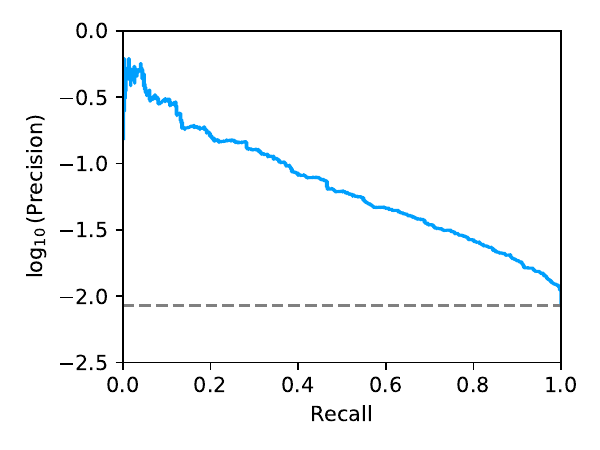}
    \caption{(Log) precision-recall. AUC: 0.10.}
    \end{subfigure}
    \begin{subfigure}[t]{0.34\linewidth}
    \centering
    \includegraphics[width=\linewidth]{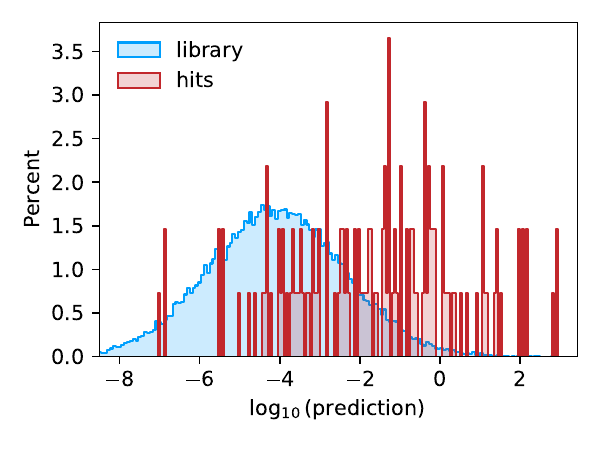}
    \caption{Predictive scores on library sequences (blue) versus heldout positive examples (red). }
    \end{subfigure}
    \begin{subfigure}[t]{0.3\linewidth}
    \centering
    \includegraphics[width=\linewidth]{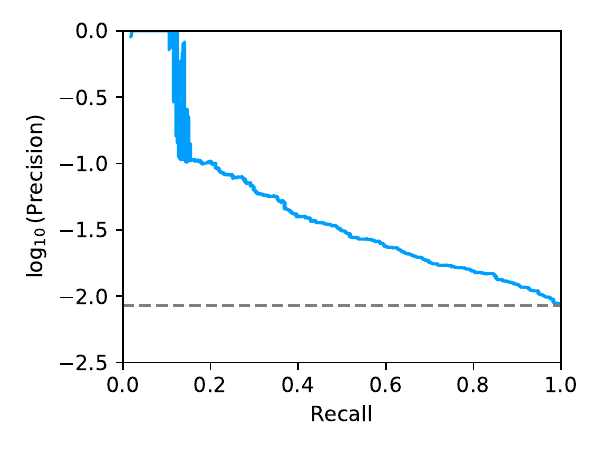}
    \caption{(Log) precision-recall using sequenced $x$ in place of computational samples from the synthesis model. AUC: 0.14.}
    \end{subfigure}
    
\caption{Predicting binding of TCR mimicking scFv CAR therapeutic candidates  to MAGE-A4, a challenging oncology target. Here we use a CNN-based architecture, rather than a transformer as in \Cref{fig:experiment}. }
    \label{fig:cnn_experiment}
\end{figure*}

\vfill

\end{document}